\documentclass[twoside,twocolumn,journal]{IEEEtran}

\usepackage{amsmath,amsfonts,epsfig,graphicx,subfigure,url,multirow,booktabs,collcell}
\usepackage{array,multirow,graphicx}
\ifCLASSINFOpdf
\else
\fi
\begin{document}
%
\title{Sequential Deep Trajectory Descriptor for Action Recognition with Three-stream CNN}
%
%
%

\author{\IEEEauthorblockN{Yemin Shi, Yonghong Tian\thanks{
			Manuscript received September 21,2016; revised July 16,2016 and September 21,2016; accepted February 04, 2017. This work was partially supported by the National Basic Research Program of China under grant 2015CB351806, the National Natural Science Foundation of China under contract No. 61390515, No. U1611461, No. 61425025, and No. 61471042, Beijing Municipal Commission of Science and Technology under contract No. Z151100000915070, and Shenzhen Peacock Plan. The associate editor coordinating the review of this manuscript and approving it for publication was Dr. Sen-Ching Samson Cheung.\newline
			\indent Corresponding author: Yonghong Tian (email: yhtian@pku.edu.cn) and Yaowei Wang (email: yaoweiwang@bit.edu.cn).\newline
			\indent Y. Shi, Y. Tian and T. Huang are with the National Engineering Laboratory for Video Technology, School of EE\&CS, Peking University, Beijing 100871, China and with Cooperative Medianet Innovation Center, China.\newline
			\indent Y. Wang is with School of Information and Electronics, Beijing Institute of Technology, Beijing, China.
		}, \textit{Senior Member, IEEE}}, Yaowei Wang, \textit{Member, IEEE}, and\\
	Tiejun Huang, \textit{Senior Member, IEEE}}

%
%

\markboth{IEEE Transactions on Multimedia,~Vol.~15, No.~9, September~2016}%
{Shi \MakeLowercase{\textit{et al.}}: Sequential Deep Trajectory Descriptor for Action Recognition with Three-stream CNN}
%



\maketitle

\begin{abstract}
Learning the spatial-temporal representation of motion information is crucial to human action recognition. Nevertheless, most of the existing features or descriptors cannot capture motion information effectively, especially for long-term motion. To address this problem, this paper proposes a long-term motion descriptor called sequential Deep Trajectory Descriptor (sDTD). Specifically, we project dense trajectories into two-dimensional planes, and subsequently a CNN-RNN network is employed to learn an effective representation for long-term motion. Unlike the popular two-stream ConvNets, the sDTD stream is introduced into a three-stream framework so as to identify actions from a video sequence. Consequently, this three-stream framework can simultaneously capture static spatial features, short-term motion and long-term motion in the video. Extensive experiments were conducted on three challenging datasets: KTH, HMDB51 and UCF101. Experimental results show that our method achieves state-of-the-art performance on the KTH and UCF101 datasets, and is comparable to the state-of-the-art methods on the HMDB51 dataset.
\end{abstract}


%

\section{Introduction}\label{sec:introduction}

\IEEEPARstart{A}{ction} recognition is growing to be a widely-used technique in many applications, such as security surveillance, automated driving, home-care nursing and video retrieval. Generally speaking, action recognition aims at identifying the actions or behaviors of one or more persons from a video sequence. An action is typically represented in some consecutive video frames rather than one isolated frame. Naturally, motion information is highly discriminative to detect, understand and recognize actions from a video. Thus how to efficiently learn the effective spatial-temporal representation of motion information is crucial to human action recognition.

In recent years, many studies (e.g., \cite{wang2011action,wang2013action,simonyan2014two,donahue2015long}) made their efforts on representing an action with low-level visual features extracted from video frames and optical flow fields, consequently achieving a significant progress. For example, SIFT \cite{lowe2004distinctive} was extended to 3D-SIFT \cite{scovanner20073} and applied to action recognition. The work \cite{poppe2010survey} extracted Histogram of Oriented Gradients (HOG) \cite{dalal2005histograms} and Histogram of Optical Flow (HOF) at each spatial-temporal interest point, and then encoded features with Bag of Features (BoF). In order to capture the motion information more effectively, Wang \textit{et al.} \cite{wang2011action} proposed a method to extract dense trajectories by sampling and tracking dense points from each frame in multiple scales. They also extracted HOG, HOF and Motion Boundary Histogram (MBH) \cite{dalal2006human} at each point. The combination of these features was shown to further boost the final performance. The improved version of dense trajectories \cite{wang2013action} also considers the camera motion estimation and then applies the BoF or Fisher vector \cite{perronnin2010improving} to derive the final representation for each video. In \cite{zhou2015learning}, dense trajectories are employed in a joint learning framework to simultaneously identify the spatial and temporal extents of the actions of interest in training videos. However, most of these methods typically could not deal with long-term action, e.g., people hovering or slow walking.

Convolutional Neural Network (CNN) has shown a great success recently and achieves state-of-the-art performance on various tasks (e.g., \cite{he2015deep,wu2015modeling,sutskever2014sequence,wang2015action, szegedy2015going,he2015delving, dahl2012context}, etc.). It has been proven empirically that the features learnt from CNN are much better than the hand-crafted features like SIFT. In order to transfer CNN from images to videos, several models \cite{wu2015modeling,shi2015learning,wu2014exploring,zha2015exploiting} have been proposed. To capture the spatial-temporal representation from a video, Ji \textit{et al.} \cite{ji20133d} extended the traditional CNN to 3D-CNN, which gets inputs from multiple channels and performs 3D convolution. However, it achieved
lower performance compared with the hand-crafted representation \cite{wang2013action}. A two-stream ConvNets approach was proposed in \cite{simonyan2014two} by incorporating spatial and motion networks and pre-training these networks on the large ImageNet dataset, consequently achieving the state-of-the-art performance. After that, Wang \textit{et al.} \cite{wang2015towards} successfully trained very deep two-stream ConvNets on the UCF101 dataset. In their two-stream ConvNets, the spatial ConvNet operates on individual frames and performs action recognition as an image classification task. Unlike the spatial ConvNet, the motion ConvNet takes several consecutive optical flow displacement fields as its input to represent the motion between video frames. Typically, it operates on a $2L$-\textit{channel} stacked optical flow images, where $L$ is the number of frames in the window, consequently modelling the short-term motion. However, the long-term dependence of frames is still ignored in two-stream ConvNets.

Unlike these pure deep models, the work \cite{wang2015action} proposed trajectory-pooled deep-convolutional descriptor (TDD), which shares the merits of both hand-crafted features and deeply-learnt features. Hasan \textit{et al.} \cite{hasan2015continuous} proposed a continuous activity learning framework for streaming videos, which intricately ties together deep hybrid feature models and active learning. To address the cross-modal video retrieval task, Pang \textit{et al.} \cite{pang2015deep} presented a multi-pathway Deep Boltzmann Machine (DBM) dealing with low-level features of various types, which learns multi-modal signals coupled with emotions and semantics. Nevertheless, none of these models have the potential for capturing long-term dependence.

In our previous work \cite{shi2015learning}, we proposed a Deep Trajectory Descriptor (DTD) for action recognition. We extracted dense trajectories from multiple consecutive frames and then projected them onto a two-dimensional plane. This resulted in a ``Trajectory Texture'' image which could effectively characterize the relative motion in these frames. Then, CNN was utilized to learn a more compact and powerful representation of dense trajectories, just like learning the appearance texture features from an image. However, when the dense trajectories overlap too much (denoted as the overwriting problem), it will be hard for DTD to learn a good representation.

More recently, the Long Short Term Memory networks (LSTMs), a special kind of Recurrent Neural Networks (RNNs), was introduced to model long-term actions. Yue-Hei \textit{et al.} \cite{yue2015beyond} and Donahue \textit{et al.} \cite{donahue2015long} proposed their own recurrent networks respectively by connecting LSTMs to CNNs. Donahue \textit{et al.} tested their model on activity recognition, image description and video description. Wu \textit{et al.} \cite{wu2015modeling} achieved the state-of-the-art performance by connecting CNNs and LSTMs under their hybrid deep learning framework. Sharma \textit{et al.} \cite{sharma2015action} introduced the attention technique into LSTMs, which learns to focus selectively on parts of the video frames and classifies videos after a few glimpses.

In these RNN-LSTM networks, however, modeling the long-range dependencies are still problematic in practice. To address this problem, several methods have been proposed recently in the field of Neural Machine Translation (NMT), which aims at translating a sentence from one language to another automatically. The work \cite{sutskever2014sequence} found that reversing the source sequence and feeding it backwards into the encoder would improve the LSTMs' performance markedly, because it shortens the path from the decoder to the relevant parts of the encoder. Zaremba \textit{et al.} \cite{zaremba2014learning} found that feeding an input sequence twice also would help a network to better memorize things. However, for action recognition, neither reversing the video nor feeding the video twice is efficient.

\begin{figure}
	\centerline{\includegraphics[width=0.4\textwidth]{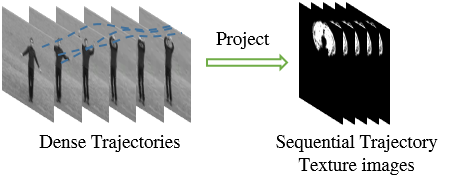}}
	\caption{Trajectories from a video are converted into a sequence of Trajectory Texture images. These images effectively characterize the relative motion in these frames and convert the raw 3D movement into multiple two-dimensional planes.}
	\label{figure:texture_image}
\end{figure}

In this paper, we propose a sequential Deep Trajectory Descriptor (sDTD) to effectively characterize long-term motion in video, consequently facilitating action recognition. A simplified version of dense trajectories is first extracted from multiple consecutive frames to represent the motion from body movement or the relative motion between the camera and objects. Then we project a set of trajectories onto a canvas, consequently resulting in a Trajectory Texture image, and trajectories from each video are converted into a sequence of Trajectory Texture images (as shown in Figure \ref{figure:texture_image}). Based on these images, a CNN can be employed to learn a macroscopical representation of motion. In order to model the dependence on the temporal domain, we consider each Trajectory Texture image as a unit and try to model them in the temporal domain with the LSTMs network. As such, videos are violently compressed over the temporal domain, making it easier for LSTMs to learn the long-term dependence.

In order to learn both spatial and motion representation, we also propose a three-stream framework. Our framework is composed of spatial stream, temporal stream and sDTD stream, which are designed to capture spatial feature, short-term feature and long-term feature respectively. The effectiveness of the proposed framework is evaluated on three public datasets: KTH, HMDB51 and UCF101. The experimental results show that our method achieves the state-of-the-art performance on the KTH and UCF101 datasets, and outperforms most methods on the HMDB51 dataset.

The remainder of this paper is organized as follows. In section \ref{sec:methodology}, the sequential Deep Trajectory Descriptor (sDTD) is introduced. The three-stream action recognition framework is presented in section \ref{sec:sDTDsystem}. The experimental
results are discussed in section \ref{sec:experiments}. Finally, section \ref{sec:conclusions} concludes this paper.

A preliminary version of this work has been published in \cite{shi2015learning}. The main extensions include
three aspects. First, we extend the Trajectory Texture image to sequential Trajectory Texture image to reduce
the pixel overwriting problem. In order to fully utilize the model pre-trained on the ImageNet dataset, we construct 3-channel Trajectory Texture images along the x and y directions of optical flows and the original direction of motion. Second, a three-stream framework is used for action recognition task, in which GoogLeNet \cite{szegedy2015going} is employed to learn sDTD, and the network is extended with LSTMs so as to model the long-term dependence.
Finally, extensive experiments are performed on more datasets so as to evaluate the effectiveness of the proposed method.

\section{Sequential Deep Trajectory Descriptor}\label{sec:methodology}

\begin{figure*}
	\centerline{\includegraphics[width=1.0\textwidth]{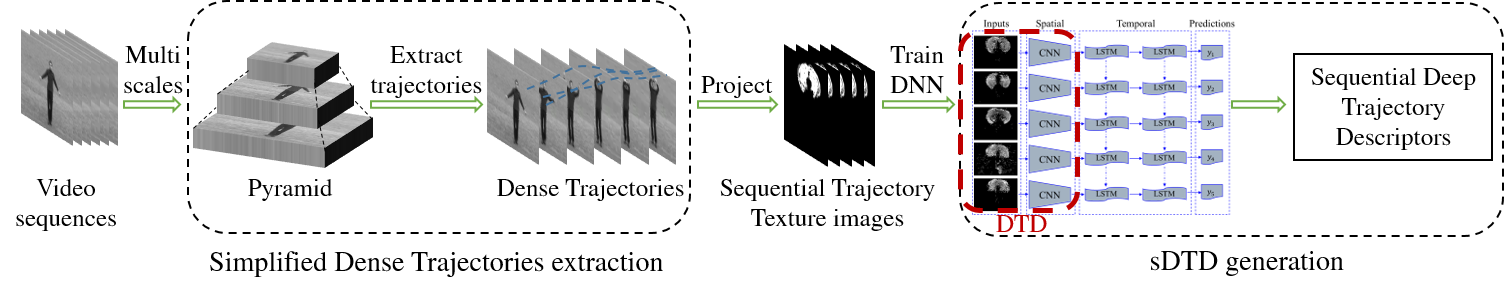}}
	\caption{\textbf{The pipeline of sDTD.} The extraction process of  sDTD is composed of three steps:
(i) extracting the simplified dense trajectories, (ii) converting these trajectories from each video into a sequence
of Trajectory Texture images, (iii) learning sDTD with a CNN-RNN network (note that the CNN part features are also called DTD \cite{shi2015learning}).}
	\label{figure:dtdfeature}
\end{figure*}

Figure \ref{figure:dtdfeature} illustrates the pipeline of our sequential Deep Trajectory Descriptor (sDTD).
Our descriptor first extracts the simplified dense trajectories and then converts these trajectories into sequential
Trajectory Texture images. Then, a deep neural network is employed to learn the descriptor for motion. This deep neural network is composed of a CNN part and a LSTM part, where the CNN part aims at learning the spatial features while the LSTM part is designed to capture the temporal features.

\subsection{The Simplified Dense Trajectories}
The initial version of dense trajectories was proposed by Wang \textit{et al.} \cite{wang2011action}.
Basically, it densely samples a set of points from multiple spatial scales on a grid with a step size of $W$ pixels, and then tracks them in multiple spatial scales. Let $(x_t,y_t)$ denote the point $p_t$ at frame $t$. Then these sampled points are tracked by median filtering in a dense optical flow field
$\omega=(u_t,v_t)$, as follows:
\begin{equation}
P_{t+1}=(x_{t+1},y_{t+1})=(x_t,y_t)+(\mathcal{M}*\omega_t)|_{(\overline{x}_t,\overline{y}_t)},
\end{equation}
where $\mathcal{M}$ is the median filter kernel, $*$ is the convolution operator, and $(\overline{x}_t,\overline{y}_t)$ is the rounded position of $(x_t,y_t)$. The maximum length of trajectories
is set as $L$ frames so as to avoid the drifting problem. A trajectory with too small or too large displacement is
also removed. Finally, a trajectory can be represented as
\begin{equation}
T_k=(\Delta d_1^k, \Delta d_2^k, ..., \Delta d_L^k),
\end{equation}
where $\Delta d_t^k$ is the displacement between $P_t$ and $P_{t+1}$. The resulting vector is normalized by the sum of the magnitudes of the displacement vectors.

The improved version of dense trajectories \cite{wang2013action} also takes the camera motion into account. It first gets the correspondence between two consecutive frames by SURF feature matching \cite{bay2006surf} and optical flow
based matching. Then, the RANSAC \cite{fischler1981random} algorithm is used to estimate the homography matrix.
Based on the homography, it rectifies the frame image to remove the camera motion and re-calculates the optical
flow. In \cite{wang2011action} and \cite{wang2013action}, HOG, HOF and MBH features are extracted along the
dense trajectories, and different features are then encoded with Bag of Feature (BoF) or Fisher vector.

Unlike the original and improved versions of dense trajectories in \cite{wang2011action} and \cite{wang2013action}, we only extract trajectories and represent them with their absolute
coordinates. That is, given a video $V$, we obtain a set of trajectories
\begin{equation}
\mathbb{T}(V)=\{T_1,T_2,...,T_K\},
\end{equation}
where $K$ is the number of trajectories, and $T_k$ denotes the $k^{th}$ trajectory
\begin{gather}
T_k=\{C_1^k, C_2^k, C_3^k, ..., C_L^k\},\\
C_l^k=(x_l^k, y_l^k, \Delta x_l^k, \Delta y_l^k),
\end{gather}
where $(x_l^k, y_l^k)$ is the spatial coordinate at $l^{th}$ time step of the trajectory $T_k$,
$(\Delta x_l^k, \Delta y_l^k)$ is the displacement between $P_l$ and $P_{l+1}$ along $x$ and $y$ axles. These
trajectories will be used to construct sequential Trajectory Texture images and finally generate the descriptor
with deep neural network, as described in the next section.

In order to compensate the camera motion, we adopt the same method as in the improved dense trajectory extraction. However, the camera motion compensation method assumes planar scenes, and complex camera motion is not considered. It is a strong assumption for real world applications and will limit our performance in 3D motion scenes. Some kinds of complex spatial analysis like 3D reconstruction will help a lot and improve the performance. This should be considered in the future works.

\subsection{The Sequential Trajectory Texture Image}\label{subsection:sTTi}
In most of existing action recognition approaches, the motion feature (e.g., HOF, MBH, temporal ConvNet feature) is vital for accurate recognition. However, extracting motion features for the actions which have long-term dependence will lead to very high computational overhead due to the processing of a large volume of video data. To address this problem, we propose a novel way to convert the motion information into two-dimensional space so that it can be efficiently processed.

Given a video $V$, we want to convert the motion in $V$ to a set of images $\mathbb{I}(V)$. Towards this end, we first extract trajectories $\mathbb{T}(V)$ from this video, which are supposed to represent all movements. For trajectory $T_k$ in $\mathbb{T}(V)$ and $I_j$ in $\mathbb{I}(V)$, we convert $T_k$ onto $I_j$ as follows:
\begin{equation}
I_j(x,y)=
\begin{cases}
\sqrt{(\Delta x_l^k)^2+(\Delta y_l^k)^2}, &if~x_l^k=x~and~y_l^k=y; \\
0, &otherwise.
\end{cases}
\label{equation:1channelproj}
\end{equation}

With Eq. (\ref{equation:1channelproj}), we are able to convert trajectories extracted from a segment of video into an image. We call it as Trajectory Texture image. However, because dense trajectories often overlap very much, there exist too many overwrites when generating the Trajectory Texture images, consequently leading to loss of recognition accuracy. To reduce the so-called overwriting problem, we further extend this equation to take the overwriting ratio into account. For image $I_j$, we define $S_n^j$ and $O_n^j$ as follows:
\begin{align}
S_n^j(x,y)=1, if~I_j(x,y)\ne 0 \label{equation:Snj},\\
O_n^j=O_{n-1}^j+\sum_{k=1}^K\sum_{l=1}^L S_n^j(x_l^k, y_l^k), \label{equation:Onj}
\end{align}
where $S_n^j(x,y)$ denotes whether it is set to a non-zero value in Trajectory Texture image $I_j$ at
position $(x,y)$ after converting all trajectories starting from the $n^{th}$ frame, and $O_n^j$ is the number of
pixels which are overwritten after the $n^{th}$ frame. When $O_n^j$ is larger than a threshold $P$, we start with a new Texture image and reset all variables.

Generally speaking, the performance of deep learning is highly related to the size of the training dataset. Labelling a large action recognition dataset and training a deep neural network model from scratch is luxurious. Thus in recent years, a popular practice to train deep networks is to utilize the pre-trained models over the ImageNet  dataset \cite{simonyan2014two, wu2015modeling, wang2015action, wang2015towards}. In order to make Eq. (\ref{equation:1channelproj}) compatible with the existing ImageNet models, we re-formulate $I_j$ to a
three-channel image:
\begin{equation}
I_j(c,x_l^k,y_l^k)=
\begin{cases}
\Delta x_l^k, &c=1; \\
\Delta y_l^k, &c=2; \\
\sqrt{(\Delta x_l^k)^2+(\Delta y_l^k)^2}, &c=3.
\end{cases}
\label{equation:3channelproj}
\end{equation}
where $I_j(c,x,y)$ is the pixel of $I_j$ in the $c^{th}$ channel at $(x,y)$, and $(x_l^k, y_l^k)$ denotes all
pixels in trajectory $T_k$. With this three-channel Trajectory Texture image, we are able to train our CNN model from the existing
ImageNet model.

\subsection{The Sequential Deep Convolutional Trajectory Descriptor}
In this subsection, we will describe how to learn the sequential Deep Trajectory Descriptor (sDTD) from a set of Trajectory Texture images using deep neural network. In principle, any kind of CNNs can be adopted. In our implementation, we tested VGG-2048 \cite{chatfield2014return} and GoogLeNet \cite{szegedy2015going}, and found that GoogLeNet was better than VGG-2048 in most cases.

\begin{figure}
	\centerline{\includegraphics[width=0.4\textwidth]{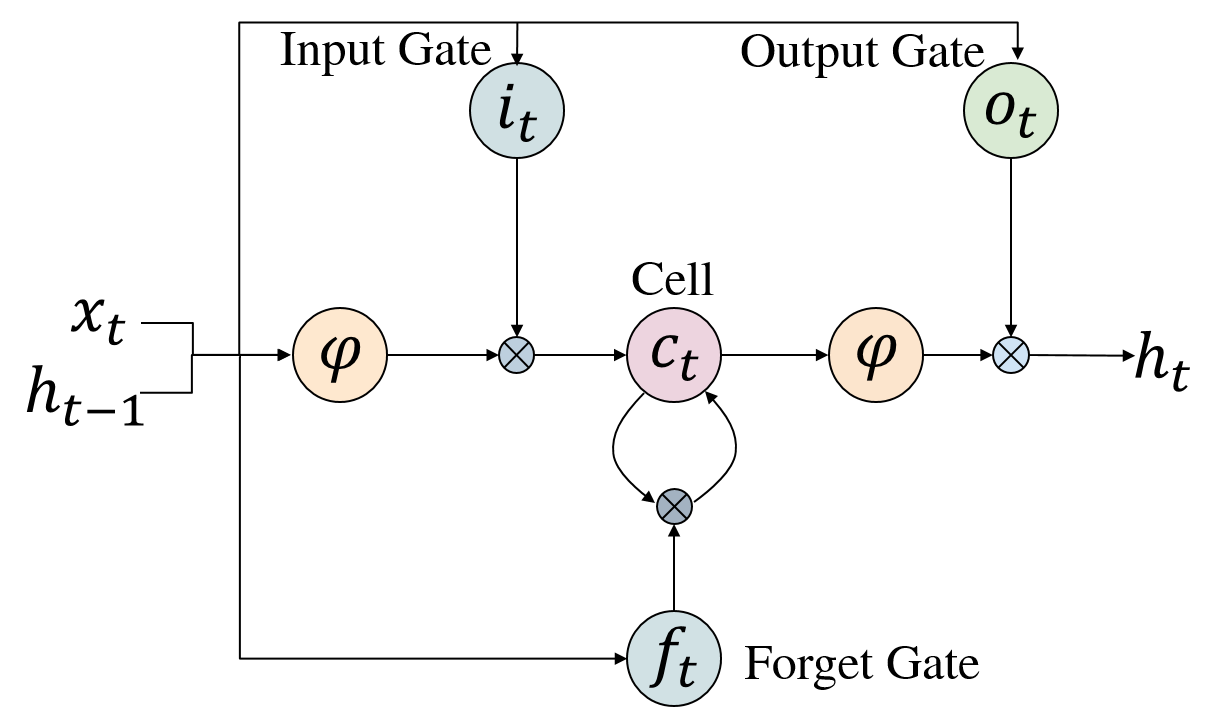}}
	\caption{Illustrating the LSTM unit used in this paper.}
	\label{figure:lstm_unit}
\end{figure}

Typically, a CNN is composed of multiple convolution layers, pooling layers and normalization layers, sometimes also includes some kinds of regularization (e.g., Dropout \cite{hinton2012improving}). In the framework of action recognition, a CNN is usually used to learn spatial features. To learn spatial-temporal features, we instead use a CNN-RNN architecture \cite{donahue2015long}. As a special kind of RNNs, LSTMs are widely used to model temporal dependency and have been successfully applied to natural language processing, speech recognition, image and video description. In our network, we will use a simplified LSTM model \cite{graves2013speech, hochreiter1997long}, which is illustrated in Figure \ref{figure:lstm_unit}.

Formally, given a sequence input as $x_t$ (where $t\in T$, $T$ is the range of time steps), an LSTM unit computes the output $h_t$ by the following equations recursively:

\begin{equation}
\begin{cases}
&i_t=\sigma(W_{xi}x_t+W_{hi}h_{t-1}+b_i), \\
&f_t=\sigma(W_{xf}x_t+W_{hf}h_{t-1}+b_f), \\
&c_t=f_t\odot c_{t-1}+i_t\odot \phi(W_{xc}x_t+W_{hc}h_{t-1}+b_c), \\
&o_t=\sigma(W_{xo}x_t+W_{ho}h_{t-1}+b_o), \\
&h_t=o_t\odot \phi(c_t),
\end{cases}
\end{equation}
where $x_t$ and $h_t$ are the input and hidden states for this LSTM unit at the $t^{th}$ time step, $i_t, f_t, c_t$, and $o_t$ are respectively the states of the input gate, forget gate, memory cell and output gate, $W_{ab}$ is the weight matrix between gate $a$ and gate $b$, $b_a$ is the bias term of gate $a$, $\sigma$ is the sigmoid nonlinearity, defined as $\sigma(x)=(1+e^{-x})^{-1}$,
which squashes real-valued inputs to a $(0,1)$ range, and $\phi$ is the hyperbolic tangent
nonlinearity, defined as $\phi(x)=\frac{e^x-e^{-x}}{e^x+e^{-x}}=2\sigma(2x)-1$.

\begin{figure}
	\centerline{\includegraphics[width=0.47\textwidth]{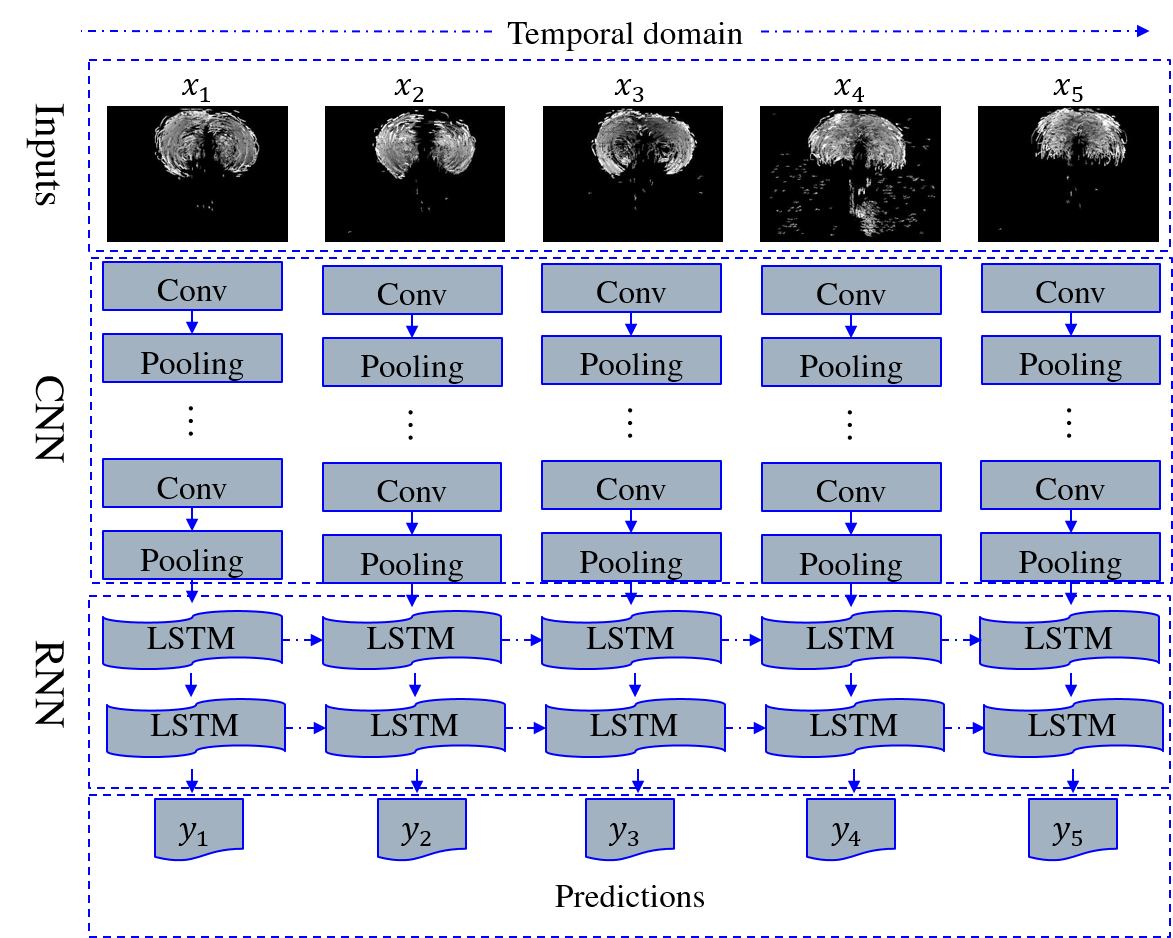}}
	\caption{The outputs of CNN are fed into LSTMs and the joint model is trained jointly. We will get a prediction for each frame and obtain the final prediction by fusing the predictions.}
	\label{figure:CNN&LSTM}
\end{figure}

To model temporal dependency, we feed the outputs of CNN into LSTMs. The joint model is shown in Figure \ref{figure:CNN&LSTM}. The whole network is composed of $4$ parts: inputs, CNNs, RNNs and predictions. The CNN part contains convolutional layers, pooling layers, and fully-connected layers. The RNN part is a multi-layer RNN network which consists of multiple LSTM layers. With this CNN-RNN architecture, we are able to fuse spatial and temporal features, and get a prediction for each frame based on its previous frames. The implementation and training details will be described in Section \ref{sec:experiments}.

\section{Action Recognition with sDTD}\label{sec:sDTDsystem}
\begin{figure*}
	\centerline{\includegraphics[width=0.9\textwidth]{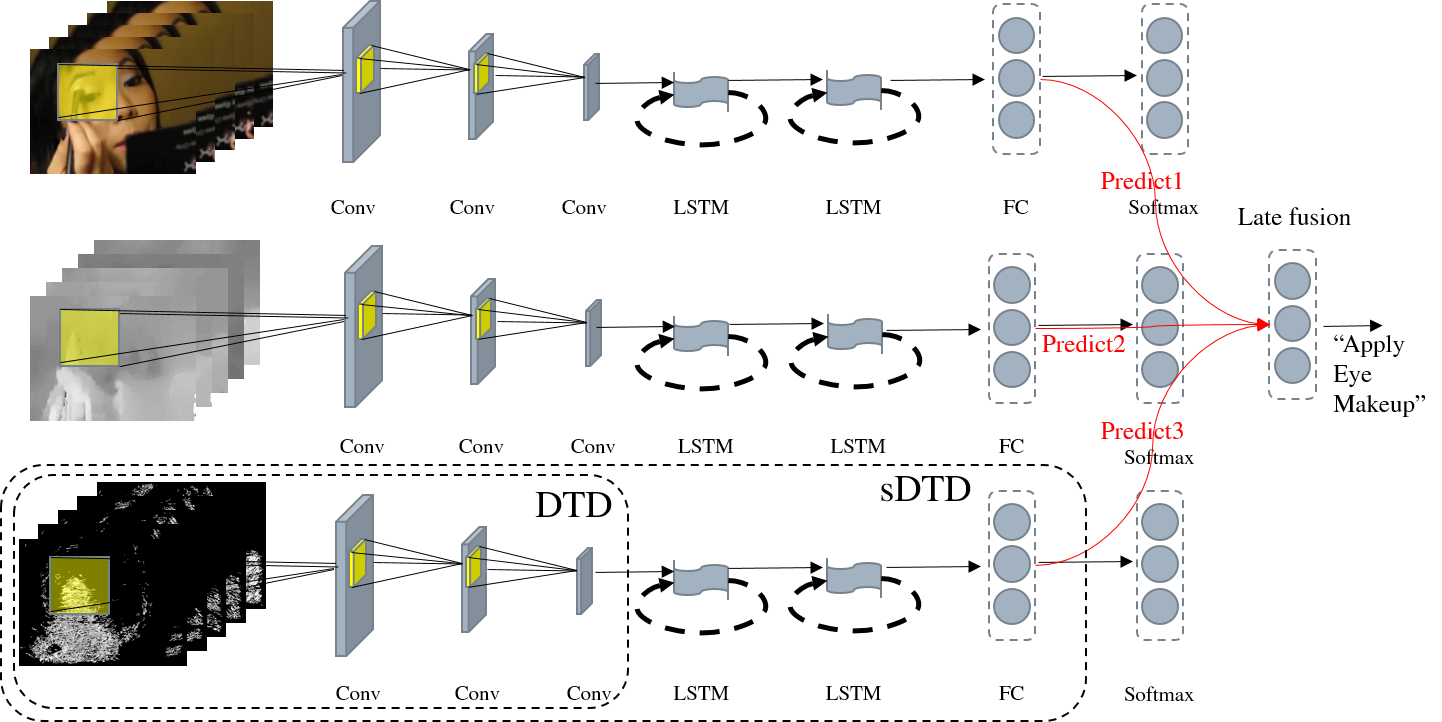}}
	\caption{Our three-stream framework for action recognition. This framework contains three streams: (i) the spatial stream for spatial feature, (ii) the temporal stream for short-term motion, (iii) the sDTD stream for long-term motion. All of the three streams have the same network architecture, which is composed of inputs, CNNs, RNNs and predictions. We do late fusion to get the final prediction for each video.}
	\label{figure:3streamCNN}
\end{figure*}

Basically, a good action recognition system should contain both spatial and temporal subsystems. In our model, we further consider the temporal subsystem as two modules: short-term temporal subsystem and long-term temporal subsystem. As a result, our three-stream framework includes spatial stream, temporal stream and sDTD stream.

The spatial stream is designed to capture static appearance features, by training on single frame images ($224\times 224\times 3$). The temporal stream takes dense optical flow fields as inputs and aims to describe the short-term motion. Unlike the two-stream ConvNets in \cite{simonyan2014two}, whose temporal stream input is volumes of stacking optical flow fields ($224\times 224\times 2F$, where $F$ is the number of stacking flows and is set to $10$), our temporal stream input is single optical flow. An optical flow field is computed from two consecutive frames and composed of vertical and horizontal flows. To make use of the optical-flow-like images, we generate the pixel at $(x,y)$ in the $c^{th}$ channel of the $t^{th}$ temporal inputs $M_t$, denoted as $M_t(c,x,y)$, as follows
\begin{equation}
M_t(c,x,y)=
\begin{cases}
u_t(x,y), &c=1 \\
v_t(x,y), &c=2 \\
\sqrt{u_t(x,y)^2+v_t(x,y)^2}, &c=3
\end{cases}
\label{equation:temporalinput}
\end{equation}
where $u_t$ and $v_t$ respectively are vertical and horizontal optical flows.

As mentioned in the previous subsection, the sDTD stream mainly focuses on modelling the long-term motion. Although both sDTD and the temporal stream utilize the optical flow fields which record the instantaneous motion state of the spatial space, they represent motion information in different manners. CNN in the temporal stream can learn the representation for short-term motion from several frames. On the contrary, sDTD samples the motion state sparsely (compared with the dense optical flow) for a video segment, and the temporal changes of spatial states result in the Trajectory Texture image, which is fed into the CNN so as to learn the representation of long-term motion. In short, the temporal stream can characterize short-term actions, while sDTD can describe long-term actions.

Finally, to get the final prediction, we apply late fusion to the three streams. Since there are three branches in GoogLeNet, thus we will obtain three predictions for each stream. In order to take full advantage of GoogLeNet, we fuse nine predictions from three streams to get the final result.

\section{Experiments}\label{sec:experiments}
This section will first introduce the detail of datasets and their corresponding evaluation schemes. Then, we describe the implementation details of our method. Finally, we report the experimental results and compare sDTD with the state-of-the-art methods.

\subsection{Datasets}
To verify the effectiveness of sDTD, we conducted experiments on three public datasets, including KTH \cite{schuldt2004recognizing}, HMDB51 \cite{kuehne2011hmdb} and UCF101 \cite{soomro2012ucf101}.

The KTH dataset contains 2,391 sequences that belong to six types of human actions by 25 subjects. These sequences are captured in four different scenarios with a homogeneous background. Following the original experimental setup, we divide the samples into the test set (9 subjects: 2, 3, 5, 6, 7, 8, 9, 10 and 22) and the training set (the remaining 16 subjects). Then we train the models on the training set and report the recognition accuracy on the test set over all classes.

The HMDB51 dataset is a large collection of realistic videos from various sources, including movies and web videos. It is composed of 6,766 video clips from 51 action categories, with each category containing at least 100 clips. Our experiments follow the original evaluation scheme, but only adopt the first training/testing split. In this split, each action class has 70 clips for training and 30 clips for testing.

The UCF101 dataset contains 13,320 video clips from 101 action classes and there are at least 100 video clips for each class. We tested our model on the first training/testing split in the experiments.

Compared with the very large dataset used for image classification, the dataset for action recognition is relatively smaller. Therefore, we pre-trained our model on the ImageNet dataset \cite{deng2009imagenet}. As UCF101 is the largest one among the three datasets, we also used it to train our three-stream model initially, and then transferred the learnt model to KTH and HMDB51.

\subsection{Implementation details}
We used the Caffe toolbox \cite{jia2014caffe} and the LSTM code in \cite{donahue2015long} to implement our model. As mentioned before,
we tested our method with VGG-2048 \cite{chatfield2014return} and GoogLeNet \cite{szegedy2015going}.

After initializing with the pre-trained ImageNet model for spatial and temporal streams, we trained our CNN-RNN network jointly. Because that sDTD will reduce the sample number for each video, it may fall in over-fitting if we train the CNN-RNN jointly in the sDTD stream. So in our implementation, we first trained a CNN model for sDTD and then added the RNN part. For KTH and HMDB51 datasets, we used the CNN-RNN model trained on the UCF101 split1 dataset and did not train the CNN separately.

The network weights were learnt using the mini-batch stochastic gradient descent with momentum (set to 0.9). The batch size for training the CNN model in the sDTD stream is 64. When training or testing a CNN-RNN model, we read 16 frames/flows/sDTDs from each video as one sample for the LSTM. For spatial and temporal streams, we read frames/flows with a stride of 5. Under this setting, we trained the CNN-RNN with a batch size of 16, which included 256 ($16\times 16$) frames/flows/sDTDs. We resized all input images to $340\times 256$, and then used the fixed-crop strategy \cite{wang2015towards} to crop a $224\times 224$ region from images or their horizontal flips. Because the 16 consecutive samples were needed in the RNN, we also forced images from the same video to crop the same region. In the test phase, we sampled 4 corners and the center from each image and its horizontal flip, and 25 samples were extracted from each video.

For spatial stream, the learning rate started from $10^{-3}$ and was divided by 10 at iteration $30K$ and $50K$, and training was stopped at $60K$ iterations. For temporal stream, we chose the TVL1 optical flow algorithm \cite{zach2007duality} and used the OpenCV GPU implementation. We discretized the optical flow fields into interval of $[0, 255]$ by a linear transformation and saved them as images. The learning rate was initially set as $10^{-3}$ and divided by 10 at iteration $80K$ and $100K$. The training was stopped at $120K$ iteration. For sDTD stream, the learning rate for training CNN started from $10^{-3}$ and decreased to $10^{-4}$ after $30K$ iterations. It was then reduced to $10^{-5}$ after $50K$ iterations and training was stopped at $60K$ iteration. When training the CNN-RNN model, the learning rate started from $10^{-3}$ and was divided by 10 every $10K$ iterations, and training was stopped at $30K$ iteration.

\subsection{Exploration experiments}

\begin{table}
    \centering
    \caption{\textnormal{Exploration of different network structures in sDTD on the UCF101 dataset. The subscripts V and G represent the VGG-2048 and GoogLeNet network structure. Here ConvNet denotes a pure CNN without LSTM layers. ST-ConvNet$_{G}$ is the fusion model of spatial ConvNet$_{G}$ and temporal ConvNet$_{G}$. Spatial and temporal streams are the first two streams in our three-stream framework, both of which have LSTM layers. DTD can be viewed as the sDTD without LSTM layers. ST-stream$_{G}$ is the fusion model of spatial stream$_{G}$ and temporal stream$_{G}$. LSTMs improve performance markedly in the three-stream framework, and GoogLeNet achieves better performance than VGG-2048.}}
    \begin{tabular}{c|c}
        \hline\hline
        model & Accuracy \\
        \hline
        Spatial ConvNet$_{G}$ & $79.0\%$ \\
        Temporal ConvNet$_{G}$ & $65.2\%$ \\
        Spatial stream$_{G}$ & $82.9\%$ \\
        Temporal stream$_{G}$ & $75.3\%$ \\
        \hline
        DTD$_{V}$ & $70.7\%$ \\
        sDTD$_{V}$ & $70.9\%$ \\
        sDTD$_{G}$ & $\textbf{71.7\%}$ \\
        \hline
        ST-ConvNet$_{G}$ & $85.5\%$ \\
        ST-stream$_{G}$ & $90.0\%$ \\
        ST-stream$_{G}$+DTD$_{V}$ & $90.9\%$ \\
        ST-stream$_{G}$+sDTD$_{V}$ & $91.8\%$ \\
        ST-stream$_{G}$+sDTD$_{G}$ & $\textbf{92.1\%}$ \\
        \hline\hline
    \end{tabular}
    \label{figure:explorationUCF101}
\end{table}

\textbf{Benefits from LSTMs.} To evaluate the contribution of LSTMs, we firstly compared the performance of CNN and CNN-RNN on the UCF101 dataset. In this experiment, we trained the CNN with Trajectory Texture images and the resulting model was named as DTD. Then we trained the CNN-RNN based on DTD. We denote the two-stream model as ST-ConvNet$_{G}$, and ST-stream$_G$ as the two-stream model with LSTM layers. The temporal ConvNet$_{G}$ operates on 20 stacked optical flow images from 11 consecutive frames. In order to show their performance in the three-stream structure, we also fused them with the spatial stream and temporal stream. The results are shown in Table \ref{figure:explorationUCF101}. We can see that the temporal ConvNet$_{G}$ gets the worst performance because no pre-trained model is available.

We can also find that the spatial and temporal streams outperform spatial and temporal ConvNets by $3.9\%$ and $10.1\%$ respectively, while the ST-stream$_G$ is $4.5\%$ better than ST-ConvNet$_G$. The remarkable improvements indicate that CNN-RNN is a better structure than the pure CNN. The DTD$_V$ and sDTD$_V$ get $70.7\%$ and $70.9\%$ respectively. It seems that LSTMs do not bring significant improvement for sDTD. However, when considering their performance in the three-stream structure, ST-stream$_{G}$+sDTD$_V$ boost from ST-stream$_{G}$+DTD$_V$ by about $1\%$. The reason may be that, LSTMs take more information into account so that sDTD can effectively encode long-term motion in the descriptor, which may be insufficient to recognize actions solely but is very informative to derive a correct prediction when fusing spatial and temporal streams. Thus, we use a CNN-RNN structure for the sDTD stream in the remainder of this section.

\textbf{Network structure.} Another important issue is the choice of network structure. We conducted an experiment on two networks: VGG-2048 and GoogLeNet and evaluated their three-stream performance on the UCF101 dataset. The results are shown in the Table \ref{figure:explorationUCF101}. We see that sDTD$_G$ is about one percentage better than sDTD$_V$. The advantage of GoogLeNet decreases to 0.3 in the three-stream framework. However, this still proves the effectiveness of GoogLeNet. Therefore, in the remainder of this section, we will use GoogLeNet to train our sDTD model and omit their subscripts.

\begin{table}
    \centering
    \caption{\textnormal{Exploration of the performance of different models on the UCF101 dataset. Our three streams are based on GoogLeNet. We compare our sDTD with iDT features \cite{wang2013action} and two-stream ConvNets \cite{simonyan2014two}. We also demonstrate the complementary properties of the three streams in the table.}}
    \begin{tabular}{c|c}
        \hline\hline
        Model & Accuracy \\
        \hline
        HOG \cite{wang2013action, wang2013lear} & $72.4\%$ \\
        HOF \cite{wang2013action, wang2013lear} & $76.0\%$ \\
        MBH \cite{wang2013action, wang2013lear} & $80.8\%$ \\
        HOF+MBH \cite{wang2013action, wang2013lear} & $82.2\%$ \\
        iDT \cite{wang2013action, wang2013lear} & $84.7\%$ \\
        \hline
        Spatial stream ConvNet \cite{simonyan2014two} & $73.0\%$ \\
        Temporal stream ConvNet \cite{simonyan2014two} & $83.7\%$ \\
        Two-stream model (fusing by SVM) \cite{simonyan2014two} & $88.0\%$ \\
        \hline
        Spatial stream & $82.9\%$ \\
        Temporal stream & $75.3\%$ \\
        sDTD & $\textbf{71.7\%}$ \\
        \hline
        ST-stream & $90.0\%$ \\
        Spatial stream+sDTD & $89.7\%$ \\
        Temporal stream+sDTD & $82.5\%$ \\
        \hline
        ST-stream+sDTD & $\textbf{92.1\%}$ \\
        \hline\hline
    \end{tabular}
    \label{figure:complementaryUCF101}
\end{table}

\textbf{Complementary properties of three streams.} Finally, in order to get a good performance on action recognition, we investigated the complementary properties of the three streams on the UCF101 dataset. The results are summarized in Table \ref{figure:complementaryUCF101}. We first fused the spatial stream and the temporal stream so as to obtain the ST-stream model. The ST-stream gets an accuracy of $90.0\%$, which is better than the two-stream model in \cite{simonyan2014two}. This improvement comes from the use of GoogLeNet and LSTMs. The GoogLeNet learns better spatial features than the shallower network in \cite{simonyan2014two}, and LSTMs perform prediction based on the previous frames. The good performance of ST-stream means that the spatial stream and the temporal stream both benefit from each other. The accuracy achieves $89.7\%$ when fusing the spatial stream with the sDTD stream, while gets $82.5\%$ when fusing the temporal stream with the sDTD stream. We can thus conclude that the spatial, temporal and sDTD streams are complementary to each other. Finally, the accuracy of $92.2\%$ for the three-stream model proves than their complementary properties can be utilized to improve the overall recognition performance.

\subsection{Evaluation of sDTD}

\begin{table}
    \centering
    \caption{\textnormal{The performance of sDTD on the KTH, HMDB51 and UCF101 datasets. Our three-stream sDTD results are obtained by fusing all branches of GoogLeNet (note that GoogLeNet has three branches) in the three streams.}}
    \begin{tabular}{c|c|c|c}
        \hline\hline
        \multirow{1}{*}{model} & \multirow{1}{*}{KTH} & \multirow{1}{*}{HMDB51} & \multirow{1}{*}{UCF101} \\
        \hline
        sDTD & $94.8\%$ & $41.1\%$ & $71.7\%$ \\
        ST-stream & $93.7\%$ & $58.4\%$ & $90.0\%$ \\
        \hline
        ST-stream+sDTD & $96.8\%$ & $63.7\%$ & $92.1\%$ \\
        Final three-stream sDTD & $\textbf{96.8}\%$ & $\textbf{65.2\%}$ & $\textbf{92.2\%}$ \\
        \hline\hline
    \end{tabular}
    \label{figure:evaluationsDTD}
\end{table}

In this subsection, we aim to evaluate the performance of our sDTD on the KTH, HMDB51 and UCF101 datasets. The experimental results are summarized in Table \ref{figure:evaluationsDTD}. We can see that, sDTD improves the ST-stream by $3.1\%$ on KTH, $5.3\%$ on HMDB51 and $2.1\%$ on UCF101.

\begin{figure*}
    \centering
    \subfigure{
        \includegraphics[width=0.14\textwidth]{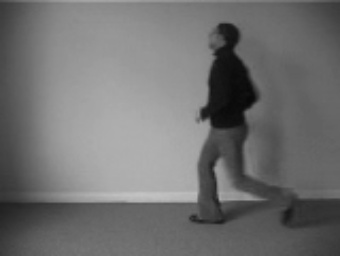}
    }
    \subfigure{
        \includegraphics[width=0.14\textwidth]{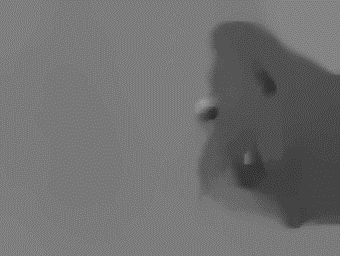}
    }
    \subfigure{
        \includegraphics[width=0.14\textwidth]{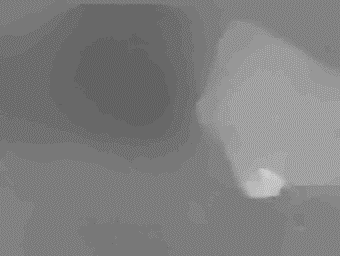}
    }
    \subfigure{
        \includegraphics[width=0.14\textwidth]{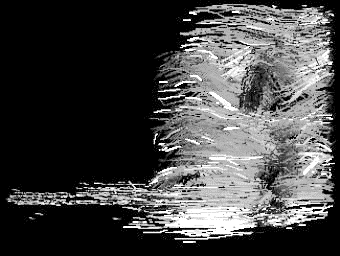}
    }
    \subfigure{
        \includegraphics[width=0.14\textwidth]{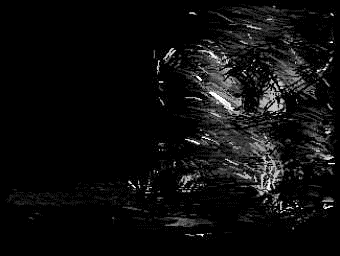}
    }
    \subfigure{
        \includegraphics[width=0.14\textwidth]{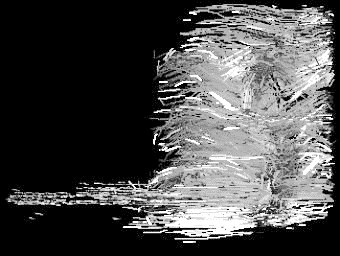}
    }
    \subfigure{
        \includegraphics[width=0.14\textwidth]{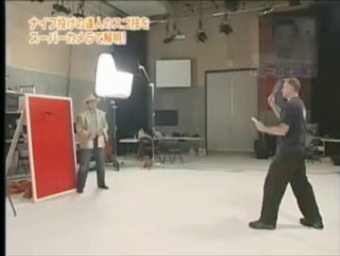}
    }
    \subfigure{
        \includegraphics[width=0.14\textwidth]{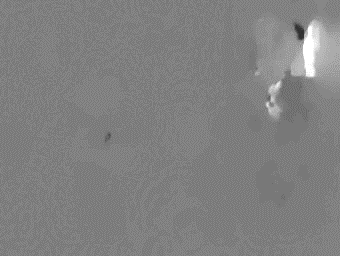}
    }
    \subfigure{
        \includegraphics[width=0.14\textwidth]{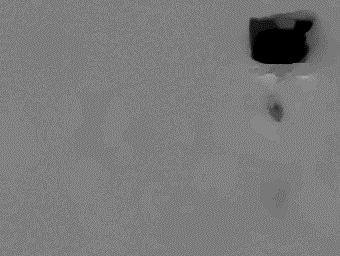}
    }
    \subfigure{
        \includegraphics[width=0.14\textwidth]{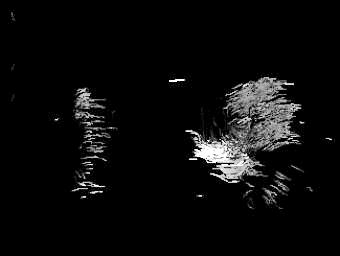}
    }
    \subfigure{
        \includegraphics[width=0.14\textwidth]{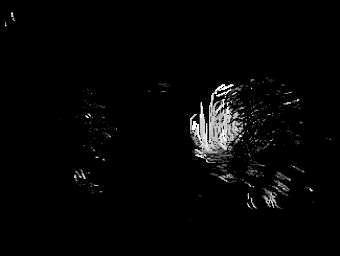}
    }
    \subfigure{
        \includegraphics[width=0.14\textwidth]{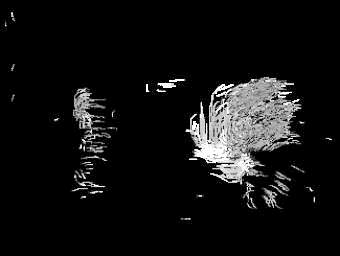}
    }
    \setcounter{subfigure}{0}
    \subfigure[RGB]{
        \includegraphics[width=0.14\textwidth]{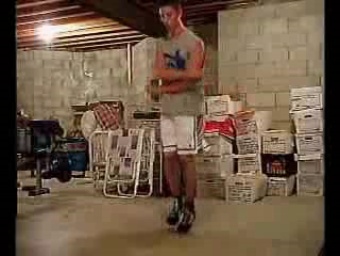}
    }
    \subfigure[Flow-x]{
        \includegraphics[width=0.14\textwidth]{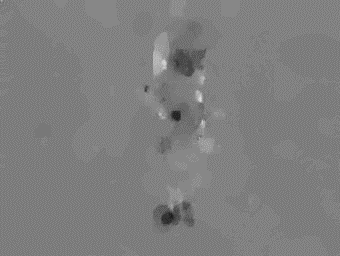}
    }
    \subfigure[Flow-y]{
        \includegraphics[width=0.14\textwidth]{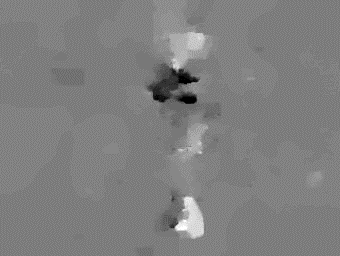}
    }
    \subfigure[TTi-x]{
        \includegraphics[width=0.14\textwidth]{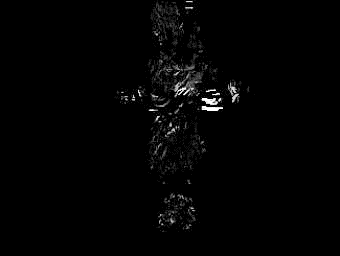}
    }
    \subfigure[TTi-y]{
        \includegraphics[width=0.14\textwidth]{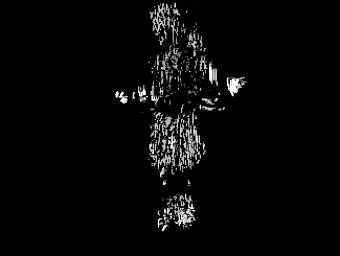}
    }
    \subfigure[TTi-m]{
        \includegraphics[width=0.14\textwidth]{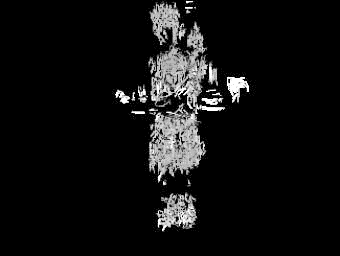}
    }
	\caption{Examples of video frames, optical flow fields, and three channels of Trajectory Texture images. Here TTi is the abbreviation of Trajectory Texture image. All images are modified to make them more visible.}
	\label{figure:examples}
\end{figure*}

Figure \ref{figure:examples} visualizes some examples of Trajectory Texture images. We can see that most background has been removed and the target object is successfully kept. And it is obvious that RGB images, optical flow fields and Trajectory Texture images can capture the visual information from different aspects, making the three streams complementary to each other.

\begin{figure*}
    \centering
    \subfigure{
        \includegraphics[width=0.3\textwidth]{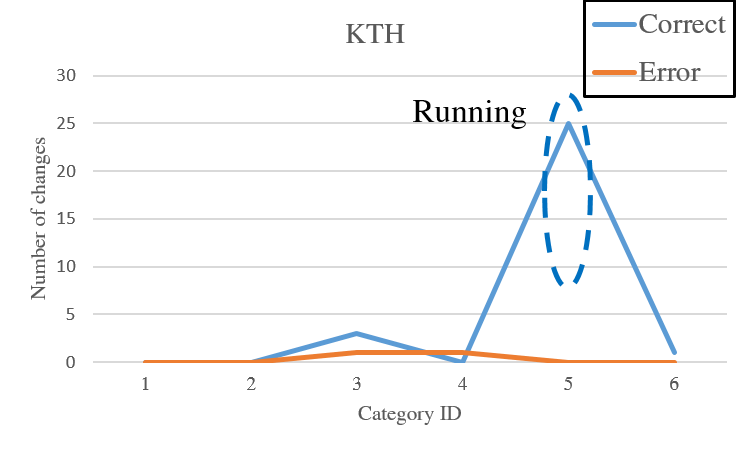}
    }
    \subfigure{
        \includegraphics[width=0.3\textwidth]{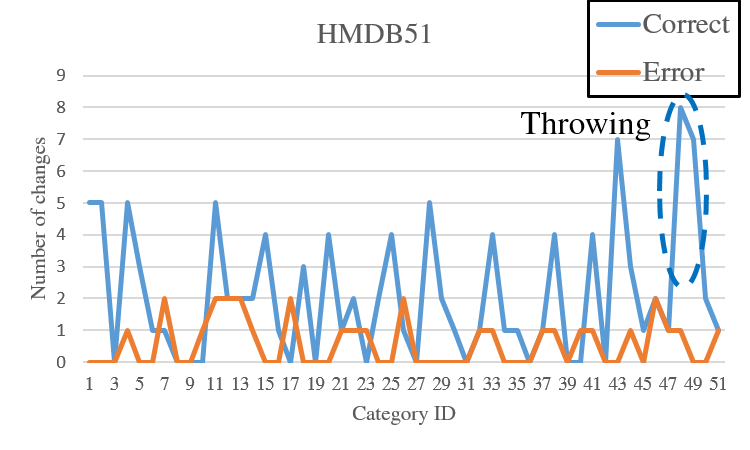}
    }
    \subfigure{
        \includegraphics[width=0.3\textwidth]{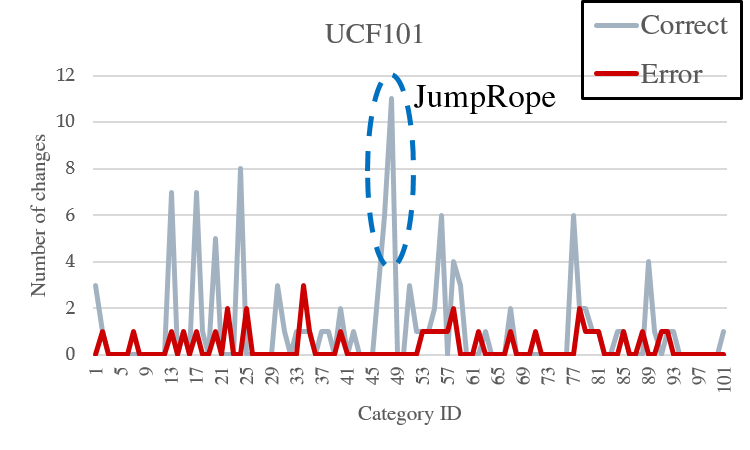}
    }
	\caption{Changes after fusing the sDTD stream into the ST-stream on the KTH, HMDB51 and UCF101 datasets. The x-axis and y-axis are category ID and the number of changed samples respectively. To get the changed samples, we simply compare the prediction results between the ST-stream and the three-streams. ``Correct'' (``Error'') means the number of videos that are predicted wrongly (correctly) before fusing the ST-stream, but correctly (wrongly) predicted with the ST-stream.}
	\label{figure:prediction_change}
\end{figure*}

In our method, each segment of a video can be converted into a few trajectory texture images (averagely 9 images in the UCF101 dataset). This can effectively reduce the number of images to be processed, consequently decreasing the computational efforts. In this way, our sDTD stream achieves a speed of 3.24 videos per second on the UCF101 dataset, which is fast enough for the real-time application.

As shown in Figure \ref{figure:prediction_change}, we count the changes after fusing the sDTD stream into the ST-stream, and sDTD brings more changes to ``Correct'' than ``Error''. A big value of ``Correct''/``Error'' means sDTD brings positive/negative effect on the ST-stream model, and zero means sDTD has no effect on the final prediction of that category. The circled parts are categories which sDTD has big effects on. As we can see, sDTD works well on \textit{Running}, \textit{Throwing} and \textit{JumpRope} classes. Specifically, before fusing sDTD, all affected \textit{Running} videos are mis-predicted as \textit{Jogging}, most affected \textit{Throwing} videos are mis-predicted as \textit{Drawing}, while most affected \textit{JumpRope} videos are mis-predicted as \textit{Basketball}, \textit{BodyWeightSquats} or \textit{SoccerJuggling}. These actions are quite similar in frames or optical flow fields when no extra objects appearing, for example, basketball. However, their long-term motion has a lot difference so sDTD is able to classify them easily.

\begin{figure}
	\centerline{\includegraphics[width=0.44\textwidth]{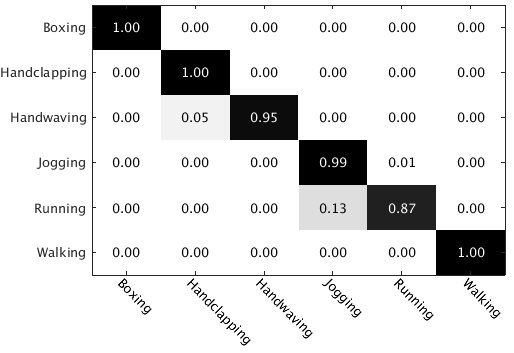}}
	\caption{Performance confusion matrix for our sDTD on the KTH dataset.}
	\label{figure:KTH_confusion_matrix}
\end{figure}
\begin{figure}
	\centerline{\includegraphics[width=0.44\textwidth]{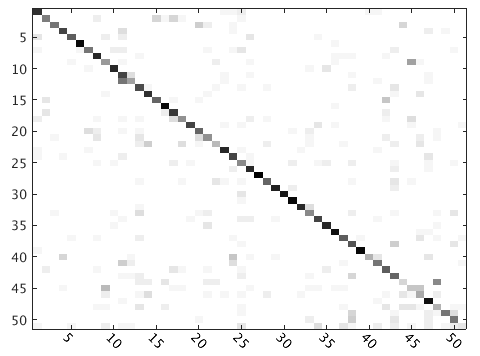}}
	\caption{Performance confusion matrix for our sDTD on the HMDB51 dataset.}
	\label{figure:HMDB_confusion_matrix}
\end{figure}
\begin{figure}
	\centerline{\includegraphics[width=0.44\textwidth]{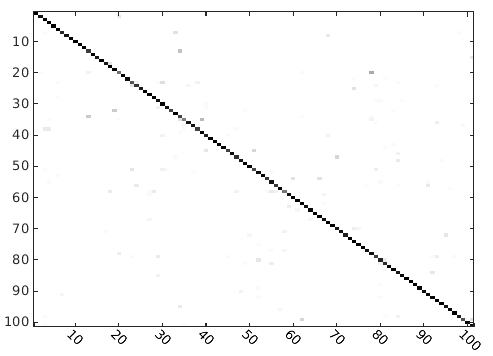}}
	\caption{Performance confusion matrix for our sDTD on the UCF101 dataset.}
	\label{figure:UCF101_confusion_matrix}
\end{figure}

The confusion matrixes for sDTD on KTH, HMDB51 and UCF101 datasets are shown in Figure \ref{figure:KTH_confusion_matrix}, \ref{figure:HMDB_confusion_matrix} and \ref{figure:UCF101_confusion_matrix}. On the KTH dataset, our method performs perfectly on \textit{Boxing}, \textit{HandClapping} and \textit{Walking} categories. The confusion matrix on the UCF101 dataset is also well diagonalized. However, the confusion matrix on the HMDB51 dataset shows that some categories are easily mis-classified, despite sDTD still performs well on most categories.

We then compare the performance of sDTD with iDT on the UCF101 dataset. They both make use of dense trajectories, but our sDTD outperforms iDT around $7.4\%$. We believe this mostly attributes to the power of deep neural network and our three-stream framework.

Finally, we compare our sDTD with DTD in \cite{shi2015learning}. The single sDTD stream achieves comparable performance with DTD+iDT on the KTH dataset, and surpasses it after fusing the spatial and temporal streams.

\subsection{Comparison to the state of the art}
\begin{table*}
    \centering
    \caption{\textnormal{Comparison of sDTD to the state-of-the-art methods.}}
    \begin{tabular}{cc|cc|cc}
        \hline\hline
        \multicolumn{2}{c|}{KTH} & \multicolumn{2}{c|}{HMDB51} & \multicolumn{2}{c}{UCF101} \\
        \hline
        HOG+HoF+BoF \cite{laptev2008learning} & $91.8\%$
            & STIP+BoF \cite{kuehne2011hmdb} & $23.0\%$
            & STIP+BoF \cite{kuehne2011hmdb} & $43.9\%$ \\
        Class-specific vocabularies \cite{kovashka2010learning} & $94.5\%$
            & Motionlets \cite{wang2013motionlets} & $42.1\%$
            & Deep Net \cite{karpathy2014large} & $63.3\%$ \\
        Hierarchical Mined \cite{gilbert2011action} & $95.7\%$
            & DT+BoF \cite{wang2013dense} & $46.6\%$
            & DT+VLAD \cite{cai2014multi} & $79.9\%$ \\
        ISA network \cite{le2011learning} & $93.9\%$
            & DT+MVSV \cite{cai2014multi} & $55.9\%$
            & DT+MVSV \cite{cai2014multi} & $83.5\%$ \\
        DT+BoF \cite{wang2011action} & $94.2\%$
            & iDT+FV \cite{wang2013action} & $57.2\%$
            & iDT+FV \cite{wang2013lear} & $85.9\%$ \\
        Dynamic coordinate system \cite{bilinski2013relative} & $94.9\%$
            & iDT+HSV \cite{peng2016bag} & $61.1\%$
            & iDT+HSV \cite{peng2016bag} & $88.0\%$ \\
        PMF+AdaBoost+Correlogram+SVM \cite{liu2013boosted} & $95.5\%$
            & PMF+AdaBoost+Correlogram+SVM \cite{liu2013boosted} & $36.5\%$
            & Hybrid deep framework \cite{wu2015modeling} & $91.3\%$ \\
        Scene Context descriptor \cite{reddy2013recognizing} & $89.8\%$
            & Two-stream model \cite{wang2015towards} & $59.4\%$
            & Two-stream model \cite{wang2015towards} & $88.0\%$ \\
        3D $\mathcal{R}$ Transform \cite{yuan20133d} & $95.5\%$
            & F$_{ST}$CN \cite{sun2015human} & $59.1\%$
            & F$_{ST}$CN \cite{sun2015human} & $88.1\%$ \\
        DTD+iDT \cite{shi2015learning} & $95.6\%$
            & TDD+FV \cite{wang2015action} & $63.2\%$
            & TDD+FV \cite{wang2015action} & $90.3\%$ \\
        - & -
            & TDD+iDT+FV \cite{wang2015action} & $\textbf{65.9\%}$
            & TDD+iDT+FV \cite{wang2015action} & $91.5\%$ \\
        - & -
            & Visual Attention \cite{sharma2015action} & $41.3\%$
            & Very deep two-stream \cite{wang2015towards} & $91.4\%$ \\
        - & -
            & - & -
            & LSTM with 30 Frame Unroll \cite{yue2015beyond} & $88.6\%$ \\
        \hline
        Three-stream sDTD & $\textbf{96.8\%}$
            & Three-stream sDTD & $\textbf{65.2\%}$
            & Three-stream sDTD & $\textbf{92.2\%}$ \\
        \hline\hline
    \end{tabular}
    \label{figure:compare_state_of_the_art}
\end{table*}

Table \ref{figure:compare_state_of_the_art} compares our recognition results with several state-of-the-art methods on the KTH, HMDB51 and UCF101 datasets. The performance of sDTD outperforms these methods on the KTH and UCF101 datasets, and outperforms most methods on the HMDB51 dataset. Unlike most of purely hand-crafted features or deep models, the work \cite{wang2015action} uses a TDD+iDT+FV framework, which takes deep feature and hand-crafted feature into one model. Our model still outperforms \cite{wang2015action} by $0.7\%$ on the UCF101 dataset. This validates the effectiveness of our sDTD.

\section{Conclusions}\label{sec:conclusions}
This paper has proposed an effective descriptor for long-term actions, sDTD. We project dense trajectories into two-dimensional planes and introduce CNN-RNN network to learn long-term motion representation. A three-stream framework is then employed to identify actions from a video sequence. Our method achieves state-of-the-art performance on the KTH and UCF101 datasets, and outperforms most of existing methods on the HMDB51 dataset.

In this paper, we have proved the importance of long-term dependence. There are also some other methods which have the potential to capture long-term dependence. Attention mechanism can be used to extract the most important frames so that we can focus on subsets of video sequences. Neural Turing Machines (NTM) \cite{graves2014neural} and Memory Networks \cite{weston2015memory} are able to help capture long-term dependence by remembering more information. These methods are important for action recognition and will be tested in our future works. In addition, we will train our model on more larger datasets like ActivityNet \cite{caba2015activitynet} or Sports-1M \cite{KarpathyCVPR14} datasets to verify the effectiveness of sDTD and improve its performance.

\ifCLASSOPTIONcaptionsoff
  \newpage
\fi



%

\bibliographystyle{./IEEEtran}
\bibliography{./IEEEabrv,./IEEEexample}

%

\begin{IEEEbiography}[{\includegraphics[width=1in,height=1.25in,clip,keepaspectratio]{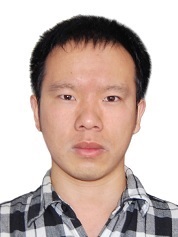}}]{Yemin Shi}
received the B.S. degree from Peking University, Beijing, China, in 2014. He is currently working toward the Ph.D. degree at the School of Electrical Engineering and Computer Science, Peking University, Beijing, China.

His research interests include machine learning, anomaly detection, action recognition and computer vision.
\end{IEEEbiography}

\begin{IEEEbiography}[{\includegraphics[width=1in,height=1.25in,clip,keepaspectratio]{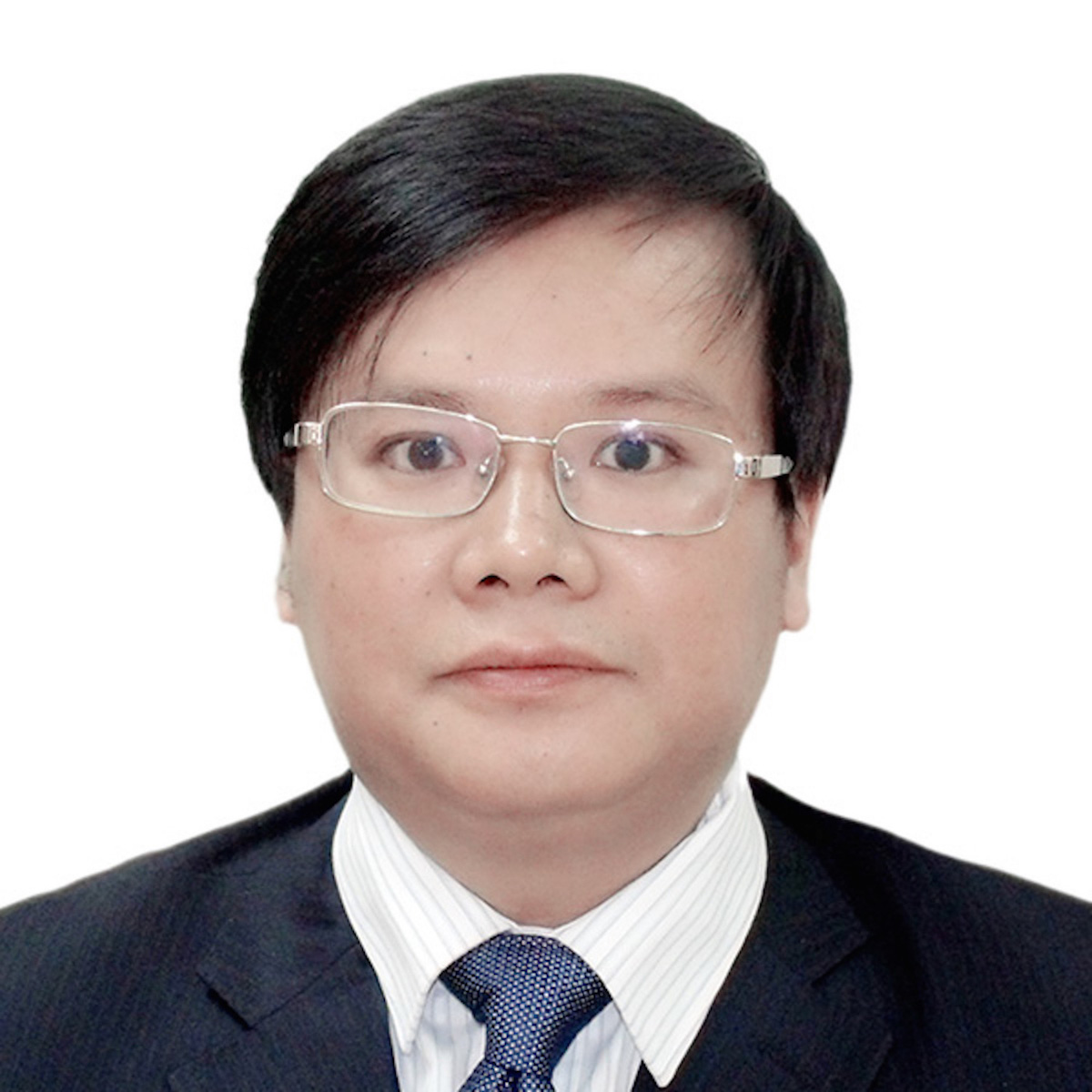}}]{Yonghong Tian}
is currently a full professor with the National Engineering Laboratory for Video Technology and the Cooperative Medianet Innovation Center, School of Electronics Engineering and Computer Science, Peking University, Beijing, China. He received the Ph.D. degree from the Institute of Computing Technology, Chinese Academy of Sciences, China, in 2005. His research interests include machine learning, computer vision, and multimedia big data. He is the author or coauthor of over 140 technical articles in refereed journals and conferences, and has owned more than 35 patents. Dr. Tian is currently an Associate Editor of IEEE Transactions on Multimedia, and International Journal of Multimedia Data Engineering and Management (IJMDEM). He has served as the Technical Program Co-chair of IEEE ICME 2015, IEEE BigMM 2015 and IEEE ISM 2015, the organizing committee member of more than ten conferences such as ACM Multimedia 2009, IEEE MMSP 2011, IEEE ISCAS 2013, IEEE ISM 2016, and the PC member of several top conferences such as CVPR, KDD, AAAI and ECCV. He was the recipient of several national and ministerial prizes in China, and obtained the 2015 EURASIP Best Paper Award for the EURASIP Journal on Image and Video Processing. His team was also ranked as one of the best performers in the TRECVID CCD/SED tasks from 2009 to 2012, PETS 2012 and the WikipediaMM task in ImageCLEF 2008. He is a senior member of IEEE and CIE, a member of ACM and CCF.
\end{IEEEbiography}

\begin{IEEEbiography}[{\includegraphics[width=1in,height=1.25in,clip,keepaspectratio]{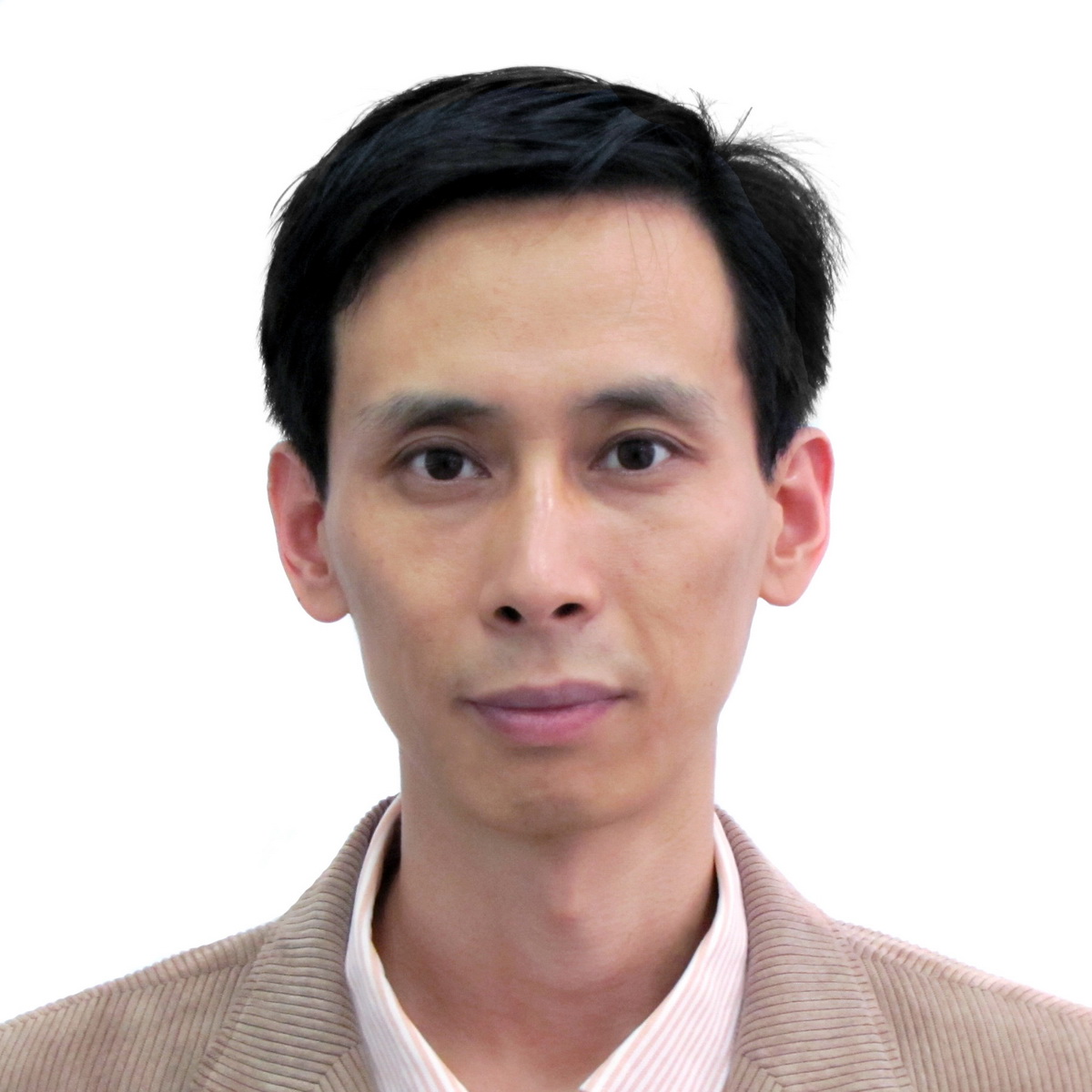}}]{Yaowei Wang,}
Ph.D., is currently an assistant professor at the Department of Electronics Engineering, Beijing Institute of Technology. He is also a guest assistant professor at the National Engineering Laboratory for Video Technology (NELVT), Peking University, China. He received Ph.D. degree on Computer Science from Graduate University of Chinese Academy of Sciences in 2005.

His research interests include machine learning, multimedia content analysis and understanding.  He is the author or coauthor of over 40 refereed journals and conference papers. His team was ranked as one of the best performers in the TRECVID CCD/SED tasks from 2009 to 2012 and PETS 2012. He is a member of IEEE and CIE.
\end{IEEEbiography}

\begin{IEEEbiography}[{\includegraphics[width=1in,height=1.25in,clip,keepaspectratio]{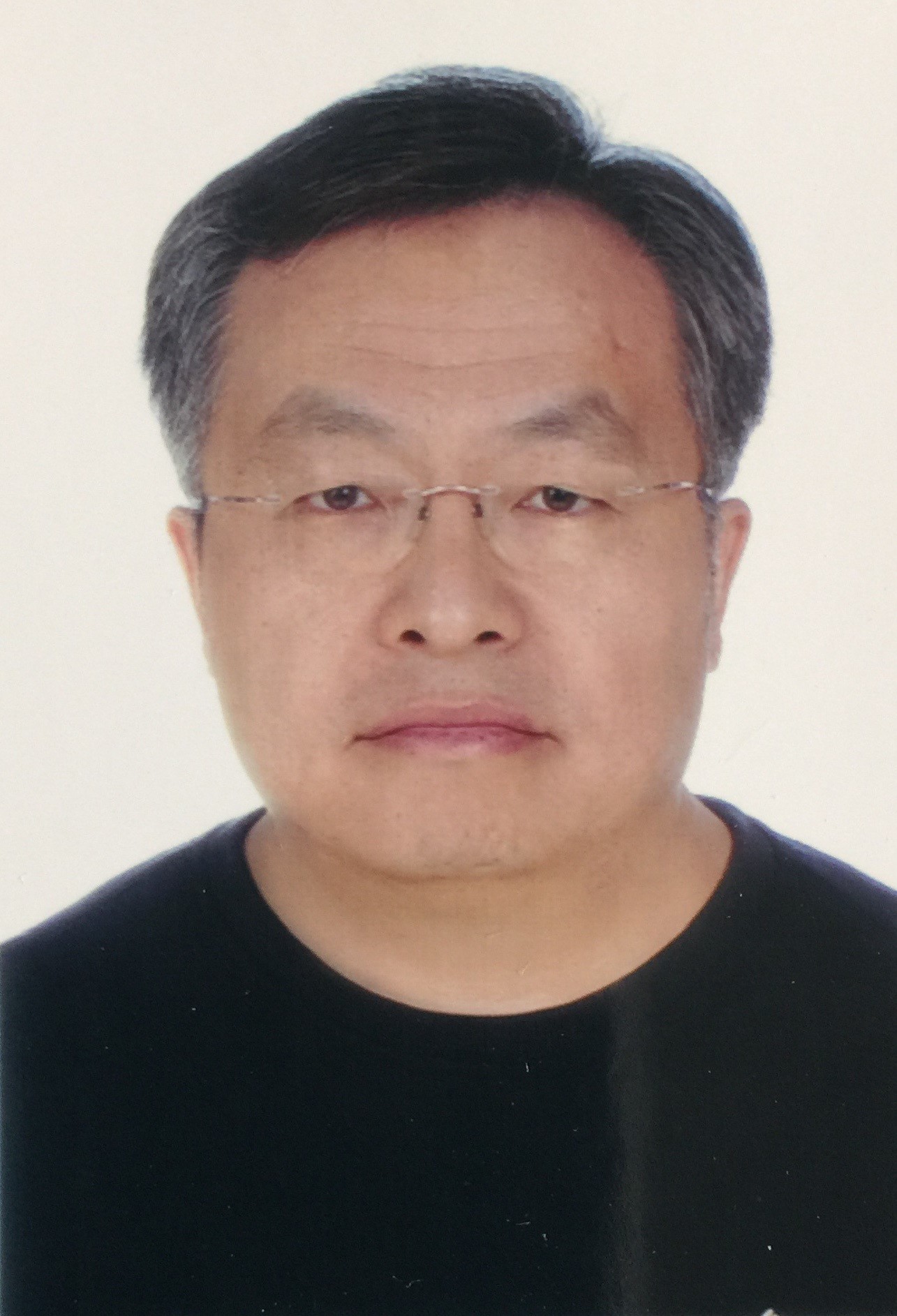}}]{Tiejun Huang,}
Ph.D, is a Professor with the School of Electronic Engineering and Computer Science, Head of Department of Computer Science, Peking University. His research areas include video coding and image understanding, especially neural coding inspired information coding theory in recent years. He received the Ph.D. degree in pattern recognition and intelligent system from the Huazhong (Central China) University of Science and Technology in 1998, and the master’s and bachelor’s degrees in computer science from the Wuhan University of Technology in 1995 and 1992, respectively. Professor Huang received the National Science Fund for Distinguished Young Scholars of China in 2014, and was awarded the Distinguished Professor of the Chang Jiang Scholars Program by the Ministry of Education in 2015. He is a member of the Board of the Chinese Institute of Electronics and the Advisory Board of IEEE Computing Now.
\end{IEEEbiography}







\end{document}